\documentclass[journal,twoside,web]{ieeecolor}
\usepackage{tmi}
\usepackage{hyperref}
\usepackage[utf8]{inputenc}
\usepackage[T1]{fontenc}
\usepackage{color,soul}
\usepackage{times}
\usepackage{amssymb} % symbols
\usepackage[nointegrals]{wasysym} % symbols
\usepackage{multicol,multirow}
\usepackage{booktabs}
\usepackage{mathtools}
\usepackage{subcaption}
\usepackage{cite}
\usepackage{amsmath,amssymb,amsfonts}
\usepackage{algorithmic}
\usepackage{graphicx}
\usepackage{textcomp}
\usepackage{adjustbox}

\begin{document}
\title{Region Guided Attention Network for Retinal Vessel Segmentation}
\author{Syed~Javed, Tariq~M.~Khan, Abdul~Qayyum, Arcot~Sowmya, Imran~Razzak
        % <-this % stops a 
\thanks{S.Javed, T.M.Khan, A. Sowmya, I. Razzak are with the School of Computer Science and Engineering, University of New South Wales, Sydney, NSW, Australia (e-mail: \{s.javed, tariq.khan, a.sowmya, imran.razzak\}@unsw.edu.au).}% <-this % stops a space
\thanks{A. Qayyum is with the National Heart and Lung Institute, Faculty of Medicine, Imperial College London, London, United Kingdom (e-mail: a.qayyum@imperial.ac.uk).}}%

\maketitle              % typeset the header of the contribution
\begin{abstract}
Retinal imaging has emerged as a promising method of addressing this challenge, taking advantage of the unique structure of the retina. The retina is an embryonic extension of the central nervous system, providing a direct in vivo window into neurological health. Recent studies have shown that specific structural changes in retinal vessels can not only serve as early indicators of various diseases but also help to understand the progression of the disease. In this work, we present a lightweight retinal vessel segmentation network based on the encoder-decoder mechanism with region-guided attention. We introduce inverse addition attention blocks with region-guided attention to focus on the foreground regions and improve the segmentation of regions of interest. To further boost the model's performance on retinal vessel segmentation, we employ a weighted dice loss. This choice is particularly effective in addressing the class imbalance issues frequently encountered in retinal vessel segmentation tasks. Dice loss penalises false positives and false negatives equally, encouraging the model to generate more accurate segmentation with improved object boundary delineation and reduced fragmentation. Extensive experiments on a benchmark dataset show better performance (0.8285, 0.8098, 0.9677, and 0.8166 recall, precision, accuracy, and F1 score, respectively) compared to state-of-the-art methods. 
\end{abstract}


\section{Introduction}
As a direct extension of the nervous system, the eye is the only part of the body where the micro-neuronal and micro-vascular systems can be viewed non-invasively, providing an accessible method for diagnosing and monitoring the effects of systemic diseases and drugs. This provides clinicians with a potential method for using the eyes to diagnose and monitor systemic disease and to assess the impact of systemic disease on eye health. Recent clinical investigations have highlighted the potential of using retinal imaging biomarkers to detect not only eye diseases (such as glaucoma and age-related macular degeneration), but also various other diseases, including hypertension, dementia, Parkinson's disease, and multiple sclerosis, in their preclinical and presymptomatic stages \cite{de2023airogs,naqvi2023glan, brahmavar2023ikd+}. In addition, it also helps diagnose conditions related to brain and heart health, which exhibit abnormal variations in the vascular structure of the retina \cite{khan2022leveraging, iqbal2022g, khan2022mkis}. Therefore, accurate segmentation of the retinal vessels makes it possible to build an automated diagnosis system, and segmentation of the retinal vessels has attracted the interest of researchers. 

Accurate segmentation of retinal vessels is hampered by the challenges posed by image features such as low contrast and imbalanced intensity, and anatomical features such as variations in thickness of the main vessels and capillaries. In addition, the presence of exudates and lesions in the image of the retinal fundus further complicates the segmentation task \cite{trinh2024sight,iqbal2022recent,de2023airogs,tang2024discriminating}. To overcome these hurdles, researchers have employed a range of supervised or unsupervised algorithms alongside computer vision techniques, aiming for accurate and automated segmentation \cite{iqbal2023robust,Khawaja2019a}. Recent advances indicate that deep learning architectures outperform other methodologies in this domain \cite{Soomro2019,khan2020region,khan2022t}. Therefore, various deep-learning strategies have contributed to the advancement of retinal vessel segmentation. 

U-Net \cite{ronneberger2015u}, originally designed for medical image segmentation, exhibits a drawback in the identification of false boundaries within retinal images alongside blood vessels. Yan et al. \cite{yan2018joint} bolstered U-Net's efficacy by implementing segment-level loss which emphasises the thickness consistency of thin vessels. Gu \textit{et al.} \cite{Gu2019CENetCE} proposed a context encoder to capture high-level features and used pre-trained ResNet blocks to improve retinal vessel segmentation. Wang et al. \cite{Wang2019a} introduced DEU-Net, which employs a fusion module function to merge a spatial path with a large kernel. This integration preserves spatial data while effectively capturing semantic details. Dulau \textit{et al.} \cite{dulau2023ensuring} developed a post-processing pipeline named VNR (Vessel Network Retrieval) to ensure a connected structure for retinal vessel networks, improving segmentation accuracy by removing misclassified pixels and reconnecting disconnected branches. Fu \textit{et al.} \cite{fu2016deepvessel} used a multiscale, multilevel convolutional neural network (CNN) to obtain a dense hierarchical representation and also incorporated a conditional random field (CRF) to model extended interactions among pixels. However, despite the efficacy of these methods, they overlook the need to optimise computational efficiency to adapt the network for use in resource-limited embedded systems. 

Recently, researchers have shown an increased interest in lightweight networks for the segmentation of general objects and medical images \cite{qayyum2023two, javed2024advancing, mazher2024self, khan2024esdmr, matloob2024lmbis, khan2024lmbf, iqbal2023ldmres, tran2022light, khan2023retinal, abbasi2023lmbis}. SegNAS3D \cite{wong2019segnas3d} introduced a framework that searches for automated network architectures for 3D image segmentation, utilising a learnable directed acyclic graph representation to optimise hyperparameters and achieve superior segmentation results with reduced computational cost and smaller network architectures compared to manual approaches. IC-Net \cite{zhao2018icnet} introduced an image cascade network that effectively reduces computation for real-time semantic segmentation and accelerates model convergence. Xception \cite{chollet2017xception} and MobileNet \cite{howard2017mobilenets} use depth-wise separable convolutions to reduce the parameter count and computational complexity, making them suitable for devices with limited computational resources. They both improve performance and efficiency in image classification and segmentation.

In this paper, we focus on a lightweight retinal vessel segmentation network based on the encoder-decoder mechanism with region-guided attention. Motivated by Xception and MobileNet, we implement depth-wise separable convolutions in the encoder and decoder blocks to minimise computational complexity and enhance model efficiency. In addition to the depth-wise convolutions, we use a reduced number of filters in both encoder and decoder blocks to increase the robustness of the model. These features make the model suitable for devices with limited computational and memory resources. We use weighted dice loss to enhance model performance on retinal vessel segmentation, as it efficiently handles class imbalance issues commonly encountered in retinal vessel segmentation tasks. By penalising false positives and false negatives equally, Dice loss encourages the model to produce more accurate segmentation with improved delineation of object boundaries and reduced fragmentation. In addition, we introduce Inverse addition Attention (IAA) blocks to focus on the foreground regions and improve the segmentation of the region of interest (ROI). The IAA blocks dramatically improve model performance. The main contributions of this work are: 
\begin{enumerate}
\item We present a lightweight region-guided segmentation network with only 40K parameters that can be deployed on devices with limited computational and memory resources.
\item We introduce region-guided inverse addition attention blocks along with weighted dice loss specifically crafted for retinal vessel segmentation that explicitly focuses on foreground regions, resulting in better segmentation of the ROI.
\item  To refine the initial segmentation maps to obtain the refined segmentation, we propose a partial decoder and use multiple attention blocks that align the high- and low-level features.
\item We have performed extensive experiments and identified the best hyperparameters for the segmentation of retinal vessels on benchmark datasets.
\end{enumerate}

\section{Related Work}
Retinal vessel segmentation is a crucial task in ophthalmic image analysis, enabling the detection and monitoring of various eye diseases such as diabetic retinopathy, glaucoma, and hypertensive retinopathy. Over the years, numerous approaches have been proposed that use both traditional image processing techniques and advanced machine learning algorithms. Earlier methods for segmentation of the retinal vessels, that relied primarily on traditional image processing, typically involved several steps including pre-processing, vessel enhancement, segmentation, and post-processing. To improve segmentation performance, some researchers used preprocessing techniques to enhance the image quality. However, advanced machine learning-based techniques can automatically learn features from large datasets, often outperforming traditional approaches. 

CNN-based techniques have exhibited promising performance in segmentation tasks and therefore have gained popularity. Uysal et al.\cite{uysal2021computer} proposed a CNN architecture in an end-to-end learning framework for medical image segmentation. They used data augmentation to enlarge the dataset synthetically for better performance, yet their work faces the challenge of dependency on the dataset and the model overfits on a majority of medical image datasets due to large model capacity and small dataset size. Oliveira et al.\cite{OLIVEIRA2018229} used a fully convolutional neural network (FCN) for the task of segmenting retinal vessels from fundus images. FCNs are particularly well suited for this task because of their ability to produce pixel-wise predictions, which is crucial to accurately delineating the thin and intricate structures of retinal vessels. Yan \textit{et al.}\cite{8476171}, proposed a three-stage FCN architecture to progressively refine predictions through multiple stages, each stage building on the output of the previous one. In the first stage, the model gives a coarse prediction, in the second stage the coarse predictions are refined by focusing on medium-level features and improving the resolution, while in the final stage, the model achieves high precision in segmentation by correcting small errors and adding more fine details. The three-stage FCN network proved to be very effective for the delineation and segmentation of retinal vessels. Guo et al. proposed the BTS-DSN model \cite{Guo2019}, which incorporates auxiliary supervision signals at multiple intermediate layers. This approach aids in facilitating gradient flow during the training process, leading to more stable convergence and improved overall segmentation performance. The network architecture includes short connections, akin to those found in residual networks. These connections are crucial for effectively propagating information across different layers, thus improving feature extraction and improving segmentation accuracy. The authors use ResNet-101 as the backbone of the BTS-DSN model, providing it with substantial capacity. However, this choice also renders the model computationally intensive and resource-demanding, which may be a consideration for practical deployment. Arsalan et al.\cite{Arsalan2019} proposed an AI-based semantic segmentation architecture tailored for the analysis of retinal images, which leverages a deep learning framework with multiple CNN layers to achieve high precision in identifying and segmenting retinal regions affected by diabetic and hypertensive retinopathy.

U-Net\cite{ronneberger2015u} achieved ground-breaking results in image segmentation tasks and since then researchers have come up with numerous variations of the popular encoder-decoder-based architecture and have improved segmentation accuracy in general object segmentation tasks, as well as on medical image segmentation. Oktay et al.\cite{oktay2018attention} added attention gates to the standard U-Net to achieve better model sensitivity and segmentation accuracy. The authors combined the attention mechanism with the U-Net architecture for the task of segmenting multiclass medical images and obtained promising results while keeping low computational complexity. Jin \textit{et al.}\cite{Jin2019} introduced the DUNet architecture, which integrates deformable convolutional networks into the U-Net framework. The architecture was designed to capture more complex vessel structures and improve segmentation accuracy by adapting the receptive fields to the shape of the retinal vessels. Traditional convolutional layers have fixed geometric structures, which can be limiting when dealing with irregular shapes of retinal vessels, while deformable convolutions address this by allowing the network to learn offsets for the convolutional kernels, enabling adaptive and flexible receptive fields that can better capture the variability in vessel shapes and sizes. Reza \textit{et al.}\cite{azad2019bi} introduced the use of Bidirectional Convolutional Long-Short-Term Memory (Bi-ConvLSTM) layers within the U-Net architecture. ConvLSTM layers are designed to handle spatial and temporal dependencies in data, making them suitable for tasks that require contextual understanding over sequences or spatially dependent structures. By using Bi-ConvLSTM, the model can capture dependencies in both forward and backward directions, enhancing its ability to model complex spatial relationships.
Wei et al.\cite{9535112} introduced Genetic U-Net, a framework that leverages genetic algorithms for the automatic design of deep neural networks specifically tailored for the segmentation of the retinal vessels. This approach aims to optimise the network architecture without extensive manual intervention. The paper applies Neural Architecture Search (NAS) using genetic algorithms to explore and identify optimal network structures. This method systematically evolves network architectures to enhance performance, demonstrating a sophisticated use of NAS in medical imaging. Although the genetic algorithm optimises the network architecture, scaling this approach to very large datasets or real-time applications might be challenging due to the inherent computational demands and the iterative nature of the search process. 
\begin{figure*}
\centering
\includegraphics[width=0.88\textwidth]{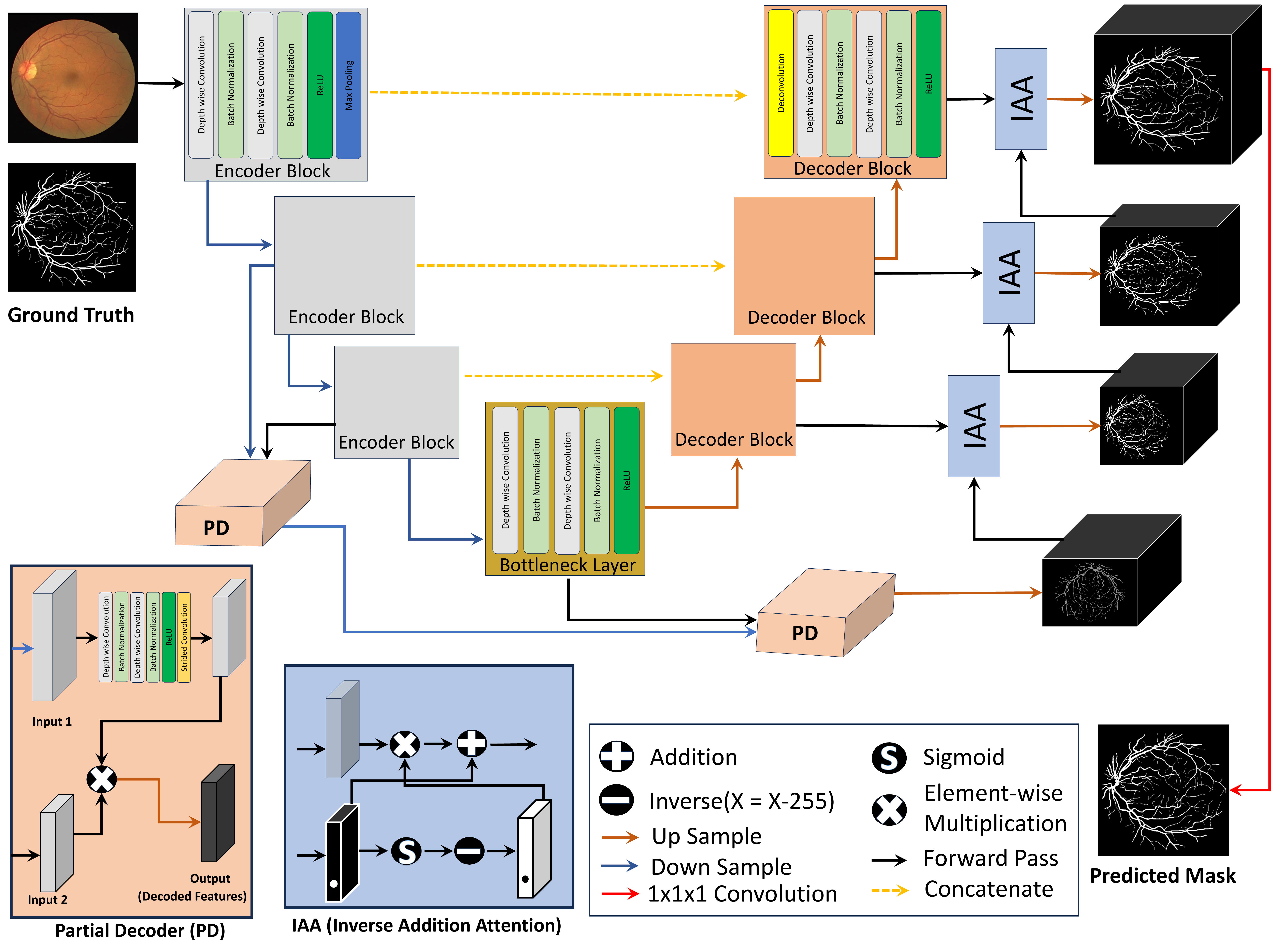}
\caption{Block diagram of the proposed methodology. The figure depicts the overall proposed methodology, detailing the Encoder, Bottleneck, and Decoder Blocks. The detailed layers and operations of the Partial Decoder and Inverse Addition Attention blocks are given in the respective sections. The figure also explains what each symbol in the flow chart means.} \label{model_diagram}
\end{figure*}

\subsection{Light-weight Models for Medical Image Segmentation}After the success of light-weight models such as Mobile-Net \cite{Howard2019MobileNet} on general object segmentation, researchers have recently been interested in developing light-weight networks for the segmentation of medical images. They have tried to minimise the size and capacity of the network, decrease the overall number of computations performed, and reduce the memory occupied by the model. Iqbal et al.\cite{iqbal2023ldmres} proposed a lightweight, compact, and efficient network called LDMRes-Net based on dual multiscale residual blocks that incorporate a multiscale feature extraction mechanism, allowing it to capture details at various levels of granularity. They reduced the number of parameters and computational complexity compared to traditional deep learning networks. The efficiency of the LDMRes-Net is enhanced by the implementation of depth-wise separable convolutions while the residual connections in the network maintain their performance. Tariq \textit{et al.} \cite{khan2024esdmr} have proposed a lightweight network for medical image segmentation. It focuses on capturing high-frequency features necessary for medical image segmentation tasks, and the implementation of expand-and-squeeze blocks makes their model robust and computationally efficient. The authors have attempted to provide a solution for applications on devices with limited computational resources. Li \textit{et al.}\cite{lightweightUnet} utilised a lightweight version of U-Net that is designed to be computationally efficient while maintaining high accuracy in the segmentation of lesions in ultrasound images. There is not much work done on the utilisation and introduction of lightweight models for the segmentation of retinal vessels. In this paper, we focus on building a lightweight model for the segmentation of retinal vessels while maintaining state-of-the-art segmentation performance.

 \section{Region Guided Attention Network}
In this work, we propose a region-guided attention network that uses the strengths of U-Net\cite{ronneberger2015u} and a region-guided attention mechanism for the segmentation of medical images. As a base model, we first modified the U-Net architecture to its lightweight version. To do so, we minimise the number of learnable parameters by reducing the number of layers and the number of filters in each layer. The motivation behind this step was to deal with the overfitting issue that occurs mainly due to the large model capacity and the small size of the medical images dataset that specifically deals with connected vessels. This choice of parameters makes the model computationally efficient while maintaining satisfactory segmentation performance and deals with weak anti-noise interference ability specifically for capillary vessels.
To boost the segmentation performance even further, we introduce an Inverse Addition Attention mechanism that forces the model to focus on the ROIs that are most relevant to the segmentation task. As shown in Figure \ref{model_diagram}, the initial segmentation map is generated by the partial decoder, and through multiple attention blocks, we refine the initial segmentation maps until we obtain the final refined segmentation map. 

\subsection{Encoder Block}
% The encoder block in our model comprises two convolutional layers, each followed by a batch normalization layer, with a ReLU activation function and followed by a non-overlapping max pooling operation. The primary goal of this block is to capture and refine relevant features before passing them to the next blocks. To enhance computational efficiency and avoid redundant operations, we employ depthwise separable convolutions. This not only speeds up model training and inference, which is critical for real-time applications but also significantly reduces the number of parameters, resulting in lower memory usage. Depthwise convolutions independently process each channel, allowing for efficient spatial feature extraction without excessive parameter overhead. To achieve a balanced integration of spatial and channel-wise features, we combine depthwise convolutions with pointwise (1x1) convolutions. This combination enables the model to effectively capture detailed and relevant features, leading to improved representation learning and overall model performance.

The encoder block in the proposed model comprises two convolutional layers, each followed by a batch normalisation layer and a ReLU activation function, and is concluded with a non-overlapping max pooling operation. The primary goal of this block is to capture and refine relevant features before passing them on to subsequent blocks. To enhance computational efficiency and avoid redundant operations, we employ depthwise separable convolutions. This approach not only accelerates model training and inference, which is critical for real-time applications but also significantly reduces the number of parameters, resulting in lower memory usage. Depthwise convolutions process each channel independently, allowing efficient spatial feature extraction without excessive parameter overhead. To achieve a balanced integration of spatial and channel-wise features, we combine depth-wise convolutions with point-wise ($1\times1$) convolutions. This combination enables the model to effectively capture detailed and relevant features, leading to improved representation learning and overall model performance. In addition, the use of batch normalisation helps stabilise and accelerate the training process by normalising the activations of each layer. This reduces internal covariate shift and enables the use of higher learning rates, further speeding up convergence. The ReLU activation function introduces nonlinearity into the model, allowing it to learn complex patterns and interactions within the data. The nonoverlapping max pooling operation reduces the spatial dimensions of the feature maps, thereby decreasing the computational load while preserving essential spatial information.

Overall, the design of the encoder block, with its separable convolutions in-depth, batch normalisation, ReLU activations, and max pooling, ensures efficient and effective feature extraction. This combination enhances the model's ability to learn rich representations from the input data, contributing to improved performance in downstream tasks.

\subsection{Decoder Block}
% A decoder block in our network comprises of a deconvolution operation that upsamples the input feature maps followed by two depth-wise separable convolutional operations similar to that in encoder blocks. A decoder block receives the input from the previous decoder block or the bottleneck layer along with features from an encoder block through a skip connection. With the deconvolution operation, the decoder will upsample the feature map from the previous block and concatenate it with the input through skip connection from the encoder block and hence retains the fine-grained spatial information for better segmentation accuracy. After the concatenation, these features are passed through convolutional layers and ReLU activation. These convolutional layers learn to refine the upsampled features and passes these features to the next decoder block.
A decoder block in the proposed network consists of a deconvolution operation that upsamples the input feature maps, followed by two depth-wise separable convolution operations similar to those in the encoder blocks. Each decoder block receives input from either the previous decoder block or the bottleneck layer, in addition to features from a corresponding encoder block via a skip connection. The deconvolution operation on the decoder up-samples the feature maps, aligning their spatial dimensions with those of the encoder features. This upsampled feature map is then concatenated with the feature map from the encoder block, preserving fine-grained spatial information critical for precise segmentation. After concatenation, the combined features undergo processing through a series of convolutional layers followed by ReLU activation. These layers refine the upsampled features, enhancing the network's ability to accurately segment the image. The output of this process is then passed on to the next decoder block, continuing the upsampling and refinement process until the original image resolution is restored. By incorporating skip connections and refining the features through convolutional layers, the decoder blocks effectively reconstruct the segmented image, maintaining spatial accuracy and detail. We also used a cascaded partial decoder to align the high- and low-level features that are then passed to the main decoders (see Figure \ref{model_diagram} partial decoder block). This addition positively impacts the segmentation result and, as shown in Table \ref{tab:Ablation}, represented by PD (Partial Decoder).
%Inspired by \cite{partial_decoder}, we use cascaded partial decoder to align high and low level features and pass it to the main decoders. This proved to positively impact our segmentation result and as shown in the table \ref{tab:Ablation}, represented by PD (Partial Decoder).

\subsection{Inverse Addition Attention Block}
% We implement inverse addition attention block to force the model to pay more attention on the foreground pixels that include the vessels. For better results we implement as many IAA blocks as the total number of decoder blocks. An IAA block receives the feature maps from the decoder block along with the segmentation map from the previous IAA block except for the first IAA block that receives one of its inputs from the partial decoder block. In the block we create a replica of the segmentation map, apply sigmoid on it, take inverse of it takes the element-wise product of the new feature map with the input features from the decoder block. Once we have this all ready, we add the actual segmentation map to the result and pass the result to the next IAA block.
We implement an Inverse Addition Attention (IAA) block to enhance the model's focus on foreground pixels, specifically those containing vessels. To achieve optimal results, we incorporate an IAA block for each decoder block in the network. Each IAA block receives feature maps from the corresponding decoder block and the segmentation map from the preceding IAA block. The exception is the first IAA block, which receives its initial input from the partial decoder block. Within each IAA block, we replicate the segmentation map, apply a sigmoid activation to it, and then compute its inverse. This inverted map is multiplied element-wise with the feature maps from the decoder block, effectively emphasising the vessel regions by suppressing the background. Following this, the actual segmentation map is added to the resulting feature map, integrating the refined attention information. The combined output is then passed to the next IAA block, continuing the process. This structured approach ensures that the model progressively refines its attention on the vessels through each stage of the decoding process, significantly improving segmentation accuracy by retaining critical spatial details and enhancing feature representation.

\subsection{Proposed Network Architecture}
To explain how the proposed network processes the input to obtain the desired segmentation map, some components of the model are first discussed. Let the model input be defined as $X_{in}$, where $X\in\mathbb{R}^{H\times W\times C}$, and $C_d^{n \times n}(*)$ be defined as a separable convolution operation in depth, where $n\times n$ is the kernel size, $\beta_{n}(*)$ is batch normalisation and $P_{max}$ be a non-overlapping max pooling operation of size $2\times 2$. Let $\Re$ be the ReLU activation and $Conv$ be the convolutional block given in Equation \ref{Eq0}. Then the first encoder block processes the input $X_{in}$ and returns two values $S_1$ and $p_1$ as in equations Eq.(\ref{Eq1}-\ref{Eq2}). 
\begin{equation}
    Conv=\Re\Bigg[\beta_{n}\Bigg(C_d^{3\times 3}\bigg(\Re\Big(\beta_{n}\big(C_d^{3\times 3}(*)\big)\Big)\bigg)\Bigg)\Bigg]
    \label{Eq0}
\end{equation}

\begin{equation}
    S_{1}=Conv(X_{in})
    \label{Eq1}
\end{equation}
\begin{equation}
    p_{1} = P_{max}(S_1)
    \label{Eq2}
\end{equation}
Here $S_1\in R^{H\times W\times 8}$ and $p_1 \in R^{\frac{H}{2}\times \frac{W}{2} \times 8}$, where $H$, $W$ and $8$ are the height, width, and channel of the feature maps respectively. Now we feed $p_1$ to the second encoder block and get $S_2 \in R^{\frac{H}{2}\times \frac{W}{2}\times16}$ and $p_2 \in R^{\frac{H}{4}\times \frac{W}{4}\times16}$ by repeating equations Eq.(\ref{Eq0}-\ref{Eq2}). We repeat the same sequence of steps for the third encoder block passing $p_2$ as the input of the layer and get $S_3 \in R^{\frac{H}{4}\times \frac{W}{4}\times 24}$ and $p_3 \in R^{\frac{H}{8}\times \frac{W}{8} \times 24}$ as the outputs. By now we are done with three encoder blocks and now we feed $p_3$ to the bottleneck layer where Eq.\ref{Eq0} is performed and we get the feature map $b\in R^{\frac{H}{8}\times \frac{W}{8}\times 32}$ which will be fed to the decoder blocks. At this point, the feature map obtained from the encoder blocks, $b$, is upsampled in the first decoder block by a deconvolution operation $D^{n\times n}(*)$ where $n$ is the kernel size. After the upsampling operation, the resultant feature map is concatenated with $S_3$ and we pass it through the convolution block Eq.\ref{Eq0}. The resultant features of the first encoder block,$dec_1$, are calculated in Eq.\ref{Eq3}.
\begin{equation}
    dec_{1}= Conv\Bigg(Concat\Big(S_3,Upsample(b)\Big)\Bigg)
    \label{Eq3}
\end{equation}
The second decoder block operates on $dec_1$ and $S_2$ as inputs and produces $dec_2$ by repeating Eq.\ref{Eq3} and likewise, the third decoder block takes $dec_2$ and $S_1$ as input and produces $dec_3$ using Eq.(\ref{Eq3}). We then use a partial decoder \cite{partial_decoder} to aggregate high-level features that have smaller spatial resolution compared to low-level features. For this purpose, we pass $dec_2$ through the convolution block (Eq.\ref{Eq0}) and get $dec_{par}$, as shown in Eq.\ref{Eq4}. 

\begin{equation}
    dec_{par} = Conv(dec_2)
    \label{Eq4}
\end{equation}
In order to refine the obtained features for more accurate segmentation, we embed an attention mechanism, namely IAA. The first attention block operates on the $dec-{par}$ and $dec_1$ to generate the first predicted segmentation mask $Pred_1$, given in Eq.\ref{Eq5}, which is later refined twice till we get the final segmentation mask.
\begin{multline}
    Pred_1 = Hadamard\Bigg(dec_1,\bigg(Inverse\Big(Sigmoid(\\dec_{par})\Big)\bigg)\Bigg) + dec_{par}
    \label{Eq5}
\end{multline}

Once we have obtained $Pred_1$, we refine it by passing it twice through the inverse addition block along $dec_2$ and $dec_3$ sequentially. $Pred_2$ is obtained by passing $Pred_1$ and $dec_2$ to Eq.\ref{Eq6} and $Pred_{final}$ is obtained by processing $Pred_2$ and $dec_3$ through Eq.\ref{Eq6}.

\begin{multline}
    Pred_{final} = Hadamard\Bigg(dec_3,\bigg(Inverse\Big(Sigmoid(\\Pred_{2})\Big)\bigg)\Bigg) + Pred_{2}
    \label{Eq6}
\end{multline}

% \end{equation}

We run $Pred_{final}$ through a $1\times 1$ convolutional layer followed by a sigmoid activation function to obtain the final predicted segmentation mask $Mask$ as given in Eq.\ref{Eq7}.

\begin{equation}
    Mask = Sigmoid\Big(Conv^{1\times 1}\big(Pred_{final}\big)\Big)
    \label{Eq7}
\end{equation}

\section{Results and Discussion}
In this section, we first describe the dataset, followed by implementation details, ablation results, and comparative analysis. 
\begin{table}
    \centering
    \caption{Datasets used in the study.}
     \setlength\tabcolsep{6pt}
    \begin{tabular}{lllll}
        \hline
        \textbf{Dataset} & \textbf{Image Resolution} & \textbf{Total} & \textbf{Training/Test Split} \\
        \hline
         DRIVE & 584$\times$565 & 40 &Train: 20, Test: 20 \\
         CHASEDB1 & 999$\times$960 & 28 &Train: 20, Test: 8 \\
         STARE & 605 $\times$700 & 20 &Train: 10, Test: 10 \\
        \hline
    \end{tabular}
    \label{tab:datasets}
\end{table}

\begin{table*}
  \centering
  \caption{Ablation study on different loss functions with basic UNet, UNet plus region guided IAA Block and with integration of cascaded partial decoder. We get the best results using UNet with an IAA block, a cascaded partial decoder, and fewer filters in the encoder and decoder blocks. *with filters (8,16,24,32)}
   \setlength\tabcolsep{6pt}
    \begin{tabular}{lccccccc}
    \toprule
    \textbf{Method} & \textbf{Loss Functions} &\textbf{Jaccard} & $\textbf{F}_\mathbf{1}$ & \textbf{Recall} & \textbf{Precision} & \textbf{Acc} & \textbf{p-value}\\
    \midrule
    UNET & IoU & 0.6602 & 0.7951 & 0.7876 & 0.8081 & 0.9648&0.493 \\
    UNET & Dice &  0.6621 & 0.7965 & 0.8042 & 0.7936 & 0.9643 &0.490 \\
    UNET & Dice + BCE  & 0.6584 & 0.7938 & 0.7850 & 0.8080 & 0.9646 & 0.481 \\
    IAA + UNET & IoU& 0.6569 & 0.7926 & 0.7766  & 0.8160 & 0.9648 & 0.484  \\
    IAA + UNET & Dice & 0.6725 & 0.8039 & 0.8055 & 0.8094 & \textit{0.9659} & 0.521\\
    IAA + UNET  & Dice + BCE & 0.6794  & 0.8087 & 0.8046 & 0.8205 & 0.9671 & 0.512\\
    IAA + PD + UNET  & Dice + BCE & 0.6805  & 0.8096 & 0.8106 & 0.8144 & 0.9666 & 0.514\\
      IAA + PD + UNET  *& Dice + BCE &\textbf{0.6903} & \textbf{0.8166} & \textbf{0.8285} & \textbf{0.8098} & \textbf{0.9677} & 0.512\\ \hline
  
    \end{tabular}
    \label{tab:Ablation}
\end{table*}

\subsection{Dataset}
We used DRIVE, CHASE\_DB1, and STARE datasets for the experiments. DRIVE contains a total of 40 images, 20 of which are used for training and 20 for testing. However, the dataset size is not suitable for deep learning purposes; therefore, we augmented the dataset. The enhancement included resizing all images from $584\times 565$ to $512\times 512$ and the application of horizontal and vertical flips along with $360$ degrees of rotation in the training images and saving the image after every degree of rotation. As a result of the augmentation, the training set size increased to $7260$ from $20$ images. Note that the test images were only resized and no other pre-processing or post-processing was performed on them. Likewise, we applied the same augmentations on the CHASE\_DB1 dataset where the 20 training images were resized and augmented while the 8 test images were resized. For the STARE dataset, we used the first $10$ images for training on which we applied horizontal and vertical flips, $360$ degrees rotation, and again horizontal and vertical flips on the rotated images. The next $10$ images were used for testing and only resized. The details of the datasets are given in Table \ref{tab:datasets}.
% We used DRIVE, STARE and CHASEDB1 datasets for our experiments the details of which are given in the table \ref{tab:datasets}.

 \subsection{Implementation Details}
All experiments were carried out on a high-performance GeForce RTX 3090 GPU, training the model for a total of 70 epochs. Initially, we attempted training for 100 epochs; however, through iterative experimentation, we observed that optimal results were consistently achieved around the 58th epoch. Therefore, we adjusted the training protocol to end at 70 epochs for subsequent experiments. The training regimen began with a learning rate of $10^{-3}$, incorporating a learning rate decay strategy throughout the training process. To optimise the model parameters, we employed the Adam optimiser with a momentum setting of $0.9$. Additionally, we implemented a learning rate reduction strategy based on plateau detection, with patience of 5 steps before triggering a reduction. In exploring various objective functions, we experimented with several commonly used loss functions, including binary cross-entropy (BCE), intersection over union (IoU), a combination of BCE and IoU, Dice loss, and combinations of Dice, BCE and IoU. Through rigorous evaluation, we determined that the weighted Dice loss consistently yielded the best results for the segmentation tasks. The selection of Dice loss as the optimal objective function is attributed to its effectiveness in addressing class imbalance, particularly prevalent in segmentation tasks such as segmentation of the retinal vessels. Its ability to provide a balanced measure of segmentation accuracy makes it suitable for a wide range of segmentation applications, ensuring robust performance across diverse datasets and tasks.

\begin{figure}
\centering
\includegraphics[width=0.48\textwidth]{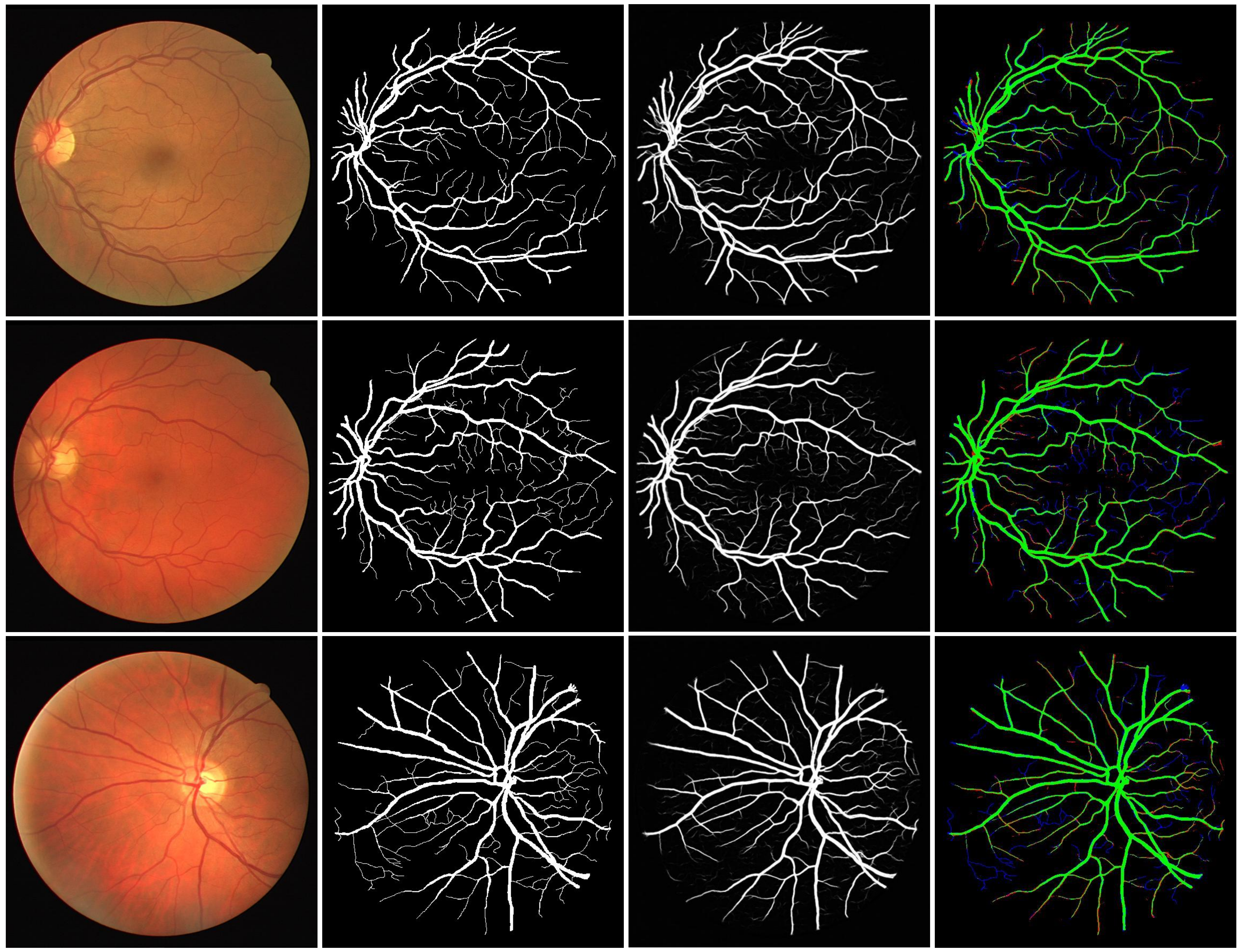}
\caption{Qualitative results of the proposed method on some sample images from the DRIVE \cite{DRIVEdata} dataset. The columns from left to right show the query image, segmentation mask (ground truth), and the mask predicted by the model and analytic mask respectively. The green pixels in the analytic mask represent the correctly segmented pixels while the red pixels are the false negatives and the blue pixels are the false positives.} \label{qresult}
\end{figure}

\begin{figure}
\centering
\includegraphics[width=0.48\textwidth]{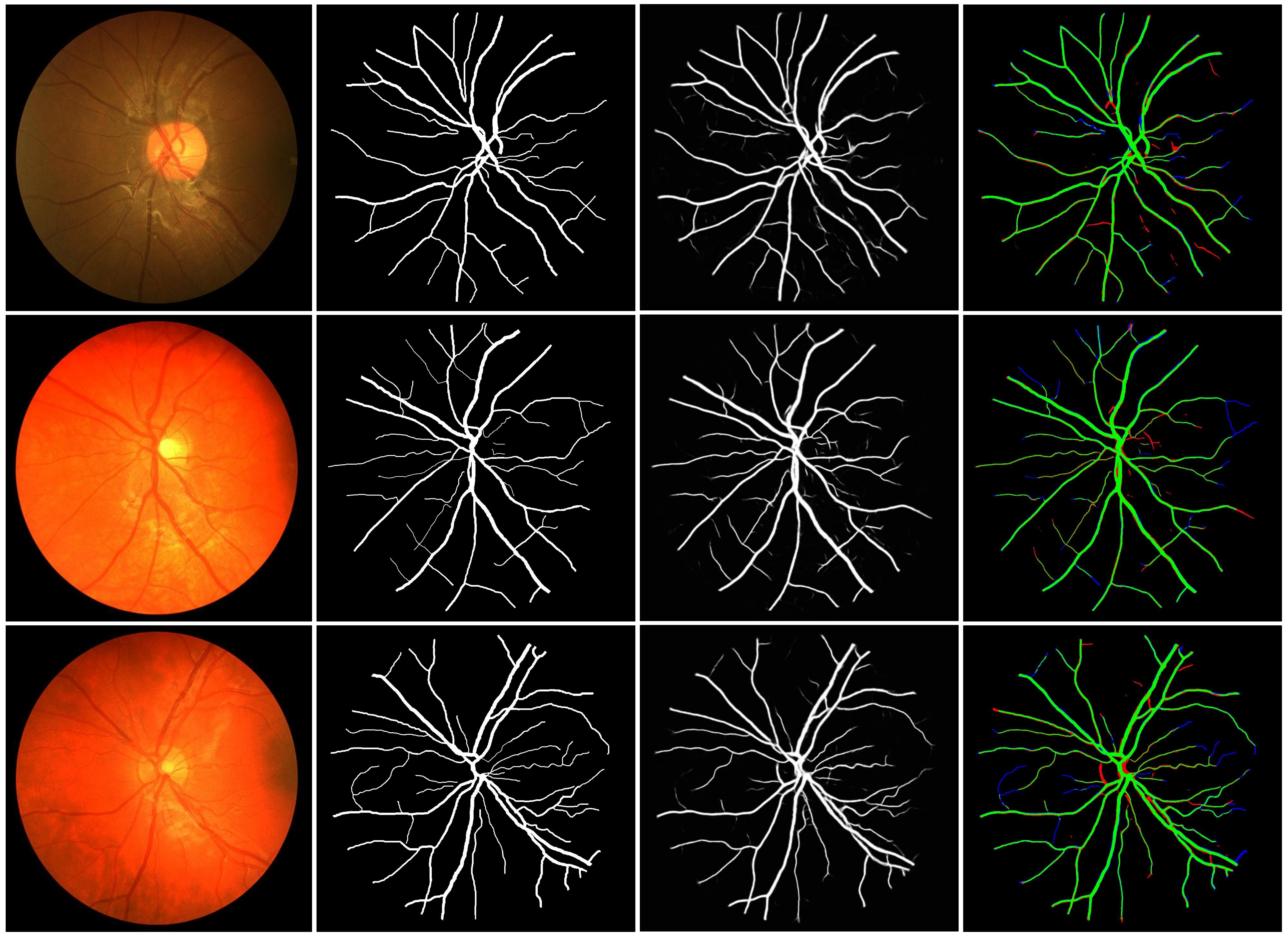}
\caption{Qualitative results of the proposed method on some sample images from the CHASEDB1 \cite{CHASEDataset} dataset. The columns from left to right show the query image, segmentation mask (ground truth), and the mask predicted by the model and analytic mask respectively. The green pixels in the analytic mask represent the correctly segmented pixels while the red pixels are the false negatives and the blue pixels are the false positives.} \label{chasevisual}
\end{figure}

\begin{table*}
  \centering
  \caption{Comparison of the proposed method with other existing works on the DRIVE \cite{DRIVEdata} dataset. The best results are in bold, and dashes indicate unknown results. Some works do not compute the $F_1$ Score for their network performance, hence there is '-' in the table.}
   \setlength\tabcolsep{6pt}
    \begin{tabular}{lccccc}
    \toprule
    \textbf{Method} & \textbf{Sensitivity} & \textbf{Specificity} & \textbf{Acc} & $\textbf{F}_\mathbf{1}$ & \textbf{Params (M)} \\
    \midrule
    VessNet \cite{Arsalan2019}  &   0.8022  & 0.9810  & 0.9655    & - & 9\\
    ERFNet \cite{Romera_2018}  & - & - & 0.9598  & 0.7652   & 2.06 \\
    UNet++ \cite{Zongwei2018} & 0.8031 & 0.9820 & 0.9533 & 0.8111& - \\
    Three-Stage FCN \cite{8476171} & 0.7631 & 0.9820 & 0.9538 & - & 20.40 \\
    FCN \cite{OLIVEIRA2018229}  & 0.8039 & 0.9804 & 0.9576& - & 0.20 \\
    M2U-Net \cite{LaibacherWJ19_CVPRW}  & - & - & 0.9630  & 0.8091   & 0.55 \\
    Vessel-Net \cite{Wu2019}  & 0.8038 & 0.9802 & 0.9578  & -   & 1.70 \\
    MobileNet-V3 \cite{howard2017mobilenets}  & 0.8250 & 0.9771 & 0.9371 & 0.6575   & 2.50 \\
   DUNet \cite{Jin2019} & 0.7963 & 0.9800  & 0.9566  & \textbf{0.8203} & 0.9 \\
    MS-NFN \cite{Wu2018}  & 0.7844 & 0.9819 & 0.9567 & -  & 0.40 \\ \hline
    Proposed Method   & \textbf{0.8285} & \textbf{0.9822} & \textbf{0.9677}& 0.8166 & \textbf{0.04}\\
    \bottomrule
    \end{tabular}
    \label{tab:DRIVE}
\end{table*}

\begin{table}
  \centering
  \setlength\tabcolsep{4pt}
    \caption{Performance comparison between the proposed method and some alternative methods on the CHASEDB1 \cite{CHASEDataset}  dataset.}
     \begin{tabular}{lccccc}
    \toprule
    \multirow{2}[4]{*}{\textbf{Method}} & \multicolumn{4}{c}{\textbf{Performance Measures in (\%)}} &\multirow{2}[4]{*}{ \textbf{Params (M)} } \\
\cmidrule{2-5}          & \textbf{Se.} & \textbf{Sp.} & \textbf{Acc} & \textbf{F1} & \\
    \midrule
    % HED \cite{xie2015holistically} & 75.16 & 98.05 & 95.97 & 97.96 & 78.15 \\
    % DeepVessel \cite{fu2016deepvessel} & 74.12 & 97.01 & 96.09 & 97.90 & - \\
    % Orlando et al. \cite{orlando2016discriminatively} & 75.65 & 96.55 & 94.67 & 94.78 & - \\
    
    Att UNet \cite{oktay2018attention} & 80.10 & 98.04 & 96.42  & 80.12 & 6.40\\
    SegNet \cite{Badrinarayanan2017} & 78.93 & 97.92 & 96.11  & 79.01 & 28.40 \\
    Wave-Net \cite{liu2022wave} & 82.83 & 98.21 & 96.64 & 83.49 & 1.5 \\
    BTS-DSN \cite{Guo2019} & 78.88 & 98.01 & 96.27 & 79.83 & 7.80\\
    DUNet \cite{Jin2019} & 77.35 & 98.01 & 96.18 &  78.83 & 0.9 \\
    % LightEyes \cite{guo2022lighteyes} & -     & -     & -     & 98.20 & - \\
    G-Net Light \cite{iqbal2022g} & 82.10 & \textbf{98.38} & 97.26 &  80.48 & 0.39 \\
    BCD-Unet \cite{azad2019bi} & 79.41 & 98.06 & 96.07 &  80.22 \\
    
    \midrule
    
    \textbf{Proposed Method} & \textbf{82.13} & 98.27 & \textbf{97.41} &  \textbf{84.59} & 0.04 \\
    \bottomrule
    \end{tabular}%
  \label{tab:CHASE}%
\end{table}%
\begin{table}
  \centering
    \caption{Performance comparison of the proposed method with a number of alternatives on the STARE \cite{STAREDataset} dataset.}
    \setlength\tabcolsep{4pt}

    \begin{tabular}{lccccc}
    \hline
    \multirow{2}[4]{*}{\textbf{Method}} & \multicolumn{4}{c}{\textbf{Performance Measures in (\%)}} &\multirow{2}[4]{*}{ \textbf{Params (M)} } \\
\cmidrule{2-5}          & \textbf{Se.} & \textbf{Sp.} & \textbf{Acc} & \textbf{F1} & \\
    \hline
    % SegNet \cite{Badrinarayanan2017} & 77.81  &  98.46  &  96.12 & 78.11 \\
    % Att UNet \cite{oktay2018attention} & 77.09  &  98.48  &  96.33  &  - \\
    % H-DenseUNet \cite{li2018h} & 78.59 & 98.42 & 96.44 & 98.47 & 82.32 \\
    Three-stage FCN \cite{yan2018three} &  77.35  &  98.57  &  96.38  &  - & 20.40\\
    BTS-DSN \cite{Guo2019} & 82.01  &  98.28  &  96.60  &  83.62 & 7.80 \\
    DUNet \cite{Jin2019} & 78.92  &  98.16  &  96.34  &  81.43 & 0.9\\
    % BCD-Unet \cite{azad2019bi} &  81.77  &  98.26  &  96.12  &  82.06 \\    
    % CC-Net \cite{Feng2020} &  80.67  &  98.16  &  96.32  &  81.36 \\
    OCE-Net \cite{OCE-NET} &  80.12  &  98.65  &  96.72  & 83.41 & 6.3\\
    Wave-Net \cite{liu2022wave} & 79.02  &  98.36  &  96.41  & 81.40 & 1.5 \\
    % LightEyes \cite{guo2022lighteyes} & -     & -     & -     & 98.29 & - \\
    G-Net Light \cite{iqbal2022g} & 81.70 &  98.53  &  97.30  &  81.78 & 0.39 \\
    LDMRes-Net\cite{iqbal2023ldmres} & 84.07 & \textbf{98.75} & 97.64 &  84.24 & 0.072 \\
    \hline
    Proposed Method&\textbf{84.64} & {98.36} & \textbf{97.91} &  \textbf{84.32} & \textbf{0.04}\\
    \hline
    \end{tabular}%
  \label{tab:STARE}%
\end{table}%

\subsection{Ablation Study}
% We conducted a comprehensive series of experiments exploring various configurations, including different loss functions and filter counts within the encoder and decoder blocks. Interestingly, our investigations revealed that instead of the standard UNET filters, we achieved optimal results with fewer filters, as outlined in Table \ref{tab:Ablation}. Our meticulous analyses showcased the superiority of the Dice loss function for retinal vessel segmentation. Furthermore, the integration of region-guided attention mechanisms significantly improved the network segmentation performance. As showcased in table \ref{tab:Ablation}, the IAA block increases segmentation performance by approximately $2.7\%$. The dice loss function effectively addresses class imbalance, leading to improved segmentation accuracy, while the reduction in the number of filters mitigates overfitting risks by limiting model capacity. This approach is particularly beneficial for handling the inherent challenges of medical image datasets, which are often small in size. \textcolor{red}{why we get better results}
We conducted a comprehensive series of experiments to explore various configurations, including different loss functions and filter counts within the encoder and decoder blocks. In particular, the investigations revealed that the use of fewer filters than the standard UNET configuration yielded optimal results, as outlined in Table \ref{tab:Ablation}. Our meticulous analyses highlighted the superiority of the Dice loss function for retinal vessel segmentation. This loss function effectively addresses class imbalance, leading to improved segmentation accuracy. Furthermore, the integration of a region-guided attention mechanism significantly improved network segmentation performance. Specifically, as shown in Table \ref{tab:Ablation}, the incorporation of the IAA block increased segmentation performance by approximately $2.7\%$. To effectively integrate global and local contexts, we used a cascaded partial decoder \cite{partial_decoder}. The global context captures the overall structure of the object, while the local context adds fine-grained details. This dual-context integration ensures a comprehensive and detailed representation of the object. The integration of PD resulted in better connected vessels and hence improved segmentation results. Finally, reducing the number of filters is particularly advantageous, as it mitigates the risk of overfitting by limiting the capacity of the model. This approach is especially beneficial for medical image datasets, which are often small in size and pose inherent challenges.

In general, the results demonstrate that carefully selecting loss functions and components of the network architecture, such as filter counts and attention mechanisms, can substantially improve segmentation performance. These findings provide valuable information for the development of more effective models for medical image analysis.
%xx

 \subsection{Comparison with Existing Works}
 
We conducted a comprehensive evaluation of our proposed model using the DRIVE \cite{DRIVEdata}, STARE \cite{STAREDataset}, and CHASEDB1 \cite{CHASEDataset} datasets, comparing its segmentation performance with existing methods. In Table \ref{tab:DRIVE} a detailed comparison is presented that shows the efficacy of the proposed model against the latest approaches on the DRIVE dataset. It is worth noting that the majority of existing work shows a remarkable gap between sensitivity and specificity, which is due to the class imbalance in terms of background and foreground pixels. The higher specificity of the models is due to the excess of background pixels in the image. However, our proposed model achieves greater sensitivity and specificity, which shows that our model segments the foreground and the background pixels better than the existing work. In addition, our model achieves the highest accuracy among existing works, showing that we capture the most retinal vessels accurately. In Tables\ref{tab:CHASE} and Table \ref{tab:STARE}, detailed comparisons of the proposed model with the state-of-the-art approaches on the CHASEDB1 and STARE datasets are presented, respectively. The results of the experiments demonstrate the notable strength and superiority of the proposed model, both in terms of segmentation performance and computational efficiency. Despite its compact size, with only $0.04$ million learnable parameters, our proposed model outperforms existing models in terms of segmentation accuracy. This underscores the efficiency and applicability of our model on devices with limited memory and computational resources while still achieving superior segmentation results compared to larger models. The proposed method achieves results superior to those of the state-of-the-art CHASEDB1 and STARE datasets. The region-guided attention block pushes our model to distinguish vessels from the background, and the low number of network parameters helps it avoid the overfitting curse commonly encountered when working with medical image datasets. Hence, our model competitively beats the state-of-the-art techniques on DRIVE, CHASE, and STARE datasets.

\subsection{Qualitative Results}
 We present some of the qualitative results obtained by our proposed method. Figure \ref{qresult} provides a qualitative analysis of our model performance on sample query images from the DRIVE dataset. Furthermore, Figure \ref{qresult} illustrates that our model accurately captures both thick and thin retinal vessels, further confirming its effectiveness in accurately segmenting the structures of the retinal vessels.
 We showcase the qualitative performance of the proposed method on the CHASEDB1 dataset \cite{CHASEDataset} in Figure \ref{chasevisual}. We distinguish the correctly labeled pixels, false positives, and false negatives in the last column of the figure. The qualitative results, too, approve the efficiency of the proposed model in segmenting retinal vessels. 
 % We performed a thorough evaluation of our proposed model on the DRIVE dataset comparing its segmentation performance with the existing works. Table \ref{tab:DRIVE} displays a comparison of our model's performance against some of the existing works on DRIVE dataset. The experiment results showcase the strength of our model and its superiority and comparability to state of the art models both in terms of efficiency and segmentation performance. If we consider the model capacity given the number of parameters, our model is very small compared to the existing models with only $0.04$ million learnable parameters. Yet, in performance our model beats the state of the art. This shows that not only our model is efficient and applicable for devices with limited memory and computational resources but also segments retinal vessels to a convenient level as compared to the existing works. Figure \ref{qresult} shows the qualitative result of our model on some sample query images from DRIVE dataset. The figure indicates that our model captures thick and thin vessels very well. 

 \section{Conclusion and Future Work}
 In this paper, we introduce a lightweight segmentation network with a significantly lower number of parameters (0.04 million), which is composed of the encoder-decoder mechanism along with partial decoder and inverse addition attention blocks for region-guided segmentation tailored specifically for retinal vessels, complemented by an in-depth ablation study focused on hyperparameter optimisation. The region-guided attention block focuses on the foreground push, whereas the cascaded partial decoder aligns the high and low level features, and hence improves the performance of the model by $2.7\%$. This comprehensive study provides researchers with valuable information, offering a solid foundation to enhance retinal vessel segmentation without the need for extensive hyperparameter optimisation. Thus, streamlining future research efforts in this domain.
 
% In this paper we not only present a lightweight segmentation network for retinal vessels but also perform an extensive ablation study on the hyper-parameters to present the optimal parameters for retinal vessel segmentation. This study gives the researchers a good head start to improve the segmentation of retinal vessels even further without having to waste time on optimizing the hyper-parameters of their network.

\bibliographystyle{elsarticle-num} 
\bibliography{egbib}





% LNCS Homepage, \url{http://www.springer.com/lncs}. Last accessed 4
% \end{thebibliography}
\end{document}